# Extending the Vocabulary of Fictional Languages using Neural Networks

**Thomas Zacharias**[*]
Electrical Engineering
Tel Aviv University
thomasz@mail.tau.ac.il

**Ashutosh Taklikar**[*]
Electrical Engineering
Tel Aviv University
ashutosht@mail.tau.ac.il

**Raja Giryes**
Electrical Engineering
Tel Aviv University
raja@tauex.tau.ac.il

**Abstract**

Fictional languages have become increasingly popular over the recent years appearing in novels, movies, TV shows, comics, and video games. While some of these fictional languages have a complete vocabulary, most do not. We propose a deep learning solution to the problem. Using style transfer and machine translation tools, we generate new words for a given target fictional language, while maintaining the style of its creator, hence extending this language vocabulary.

## Introduction

Languages can be broadly classified into natural and constructed languages. Natural languages are languages that evolved over time within a community under some cultural framework. These are constantly evolving through use and repetition without conscious planning or premeditation. Constructed languages, in contrast, are purposefully designed and are a result of a controlled intervention and language planning. Constructed languages are used for various tasks including human communication (e.g., international auxiliary language and code), giving fiction or an associated constructed setting an added layer of realism, experimentation in the fields of linguistics or cognitive science, and artistic creation (Cheyne, 2008; Sanders, 2016).

Fictional languages are constructed languages designed for a particular fictional setting - a book, movie, television show, or video game. The only native speakers of these fictional languages are the fictional characters that have been created for the fictional setting. That said, there is a large demand for fictional languages in the entertainment industry with many of these languages being used in western blockbusters, video games, comic books, and novels. Quenya and Sindarin in The Lord of the Rings, Na'vi in Avatar, Dothraki in Game of Thrones, and Klingon in Star Trek are a few examples of fictional languages with complete vocabulary and grammatical rules that anyone can learn.

Some of these languages, e.g., Na'vi and Dothraki, are constructed by professional linguists called conlangers. Yet, there are plenty of fictional works where the artists are unable to employ conlangers and need to design a language on their own. This is particularly true for fictional books, which are typically the creation of a single person. While there are exceptional authors like J.R.R. Tolkein who created multiple complete fictional languages such as Common Eldarin, Quenya, and Sindarin for The Lord of the Rings, most authors usually create a limited dictionary for their fictional language consisting of a few hundred words with their translations to English. Such languages are incomplete in two broad ways: (i) they do not have a complete vocabulary; and (ii) their grammatical structure and rules are not well defined.

Here, we propose neural network based techniques to extend the vocabulary of fictional languages, i.e., we take a limited dictionary of a few hundred words and their translations and train the networks to extrapolate the vocabulary of the language while maintaining the style of the creator. To the best of our knowledge, neural networks have not been used for such an application until now. We investigate three main architectures to address this problem: a simple Recurrent Neural Network (RNN) as a naïve solution, a Seq2Seq network, a transformer based model, and a style transfer network.

We show the validity of the proposed words by working with natural languages. We demonstrate the creation of new words in an existing natural language (English) and show that our strategy is capable of generating new words in the natural language that match preexisting words that the network has never seen. We also investigate the network performance on multiple natural languages, conduct a quantitative and qualitative analysis of the network performances, investigate linguistic connections, and provide examples of words generated by our network for a fictional language.

---
[*]T.Z. and A.T. contributed equally to this work.

While this work focuses on extending fictional vocabularies, the same technique can be used to help low-resource and endangered natural languages whose vocabularies have been shrinking over time.

**Related work**

**Language modeling** Language modeling using deep learning is a field of extensive research. Among the different architectures in this area, Recurrent Neural Networks (RNNs) and their variants are very popular (Mikolov et al., 2011). They have successfully produced results that, at first glance, look like Shakespearean work, real baby names, or even real code (Karpathy, 2015). Language modeling is appealing as networks are forced to learn about existence of words, sentences structure, and other grammatical constructs.

We focus on techniques that work on the character level. Such methods have been used to achieve interesting results in areas of question answering (Kenter et al., 2018), sentiment analysis (Radford et al., 2017) and classification (Zhang et al., 2015; Liu et al., 2017). The efficiency of such character-level neural networks for different text lengths has also been studied (Prusa and Khoshgoftaar, 2017). In our work, the first presented approach involves training a character level RNN model to learn the vocabulary of the fictional language and use this trained network to generate new words belonging to that fictional language.

**Machine Translation** Machine translation tools use RNNs (Rumelhart et al., 1985), LSTMs (Hochreiter and Schmidhuber, 1997), Seq2Seq (Sutskever et al., 2014) methods, and attention based transformer models (Vaswani et al., 2017) to translate languages. We propose using machine translation networks as vocabulary extension tools and demonstrate the same using a Seq2Seq model.

**Style transfer** For style transfer in computer vision (Liu and Tuzel, 2016; Luan et al., 2017; Zhu et al., 2017), the neural network is trained to extract content and style features from two images and then synthesize a new image by combining the content features of one image with the style of the other (Gatys et al., 2016).

Style transfer in text (Shen et al., 2017; Zhao et al., 2018) is based on a similar idea: A particular text is given as an input to a trained network and the network tries to generate new text with a different style. Even though the fundamental principle is the same, progress of style transfer in language has lagged behind computer vision mainly because of lack of parallel data and reliable evaluation metrics. While there are works implementing style transfer using parallel data (Jhamtani et al., 2017; Carlson et al., 2018; Wang et al., 2019), the small size of the available parallel corpora has made the non-parallel data approach more attractive. Language style transfer methods based on non-parallel data have been used to successfully demonstrate controlled generation of text, sentiment modification, word substitution decipherment, word order recovery, sentence rewriting, etc. (Hu et al., 2017; Shen et al., 2017; Zhao et al., 2018).

This work proposes learning the style of the limited fictional vocabulary which would allow us to generate new content (vocabulary) following the same style used by its creator. To the best of our knowledge, language style transfer has only been used on the sentence level and not for novel words creation.

**Methods**

Using RNN, Seq2Seq, and style transfer techniques we demonstrate the successful extrapolation of fictional language vocabularies. The first is a naïve approach that trains a character level RNN to learn the probability distribution of the embeddings of the characters in the fictional language and use this to generate new fictional words. The RNN model acts as a baseline and mainly emphasizes the need for a more sophisticated approach. The second approach involves using Seq2Seq networks which trains a network on parallel word translations where sequence of characters are fed as an input to the model. The final method uses style transfer methodology that trains a network to learn the underlying style of the fictional language and then use cross-alignment to generate new fictional words (Shen et al., 2017).

**Word generator using an RNN baseline.** We trained a character level RNN and analyzed its outputs. The network was trained on all the words available in the target language. The trained RNN was then given the first character of the desired new target word and was required to generate an output. For making a dictionary-like model an input source word must be mapped to a character in the target language. This initial character of the target language was based on a combination of the characters of the source word to be translated (see illustration in Fig. 1). We used the mean of the character embeddings of the source word, but any other indicator could be used.

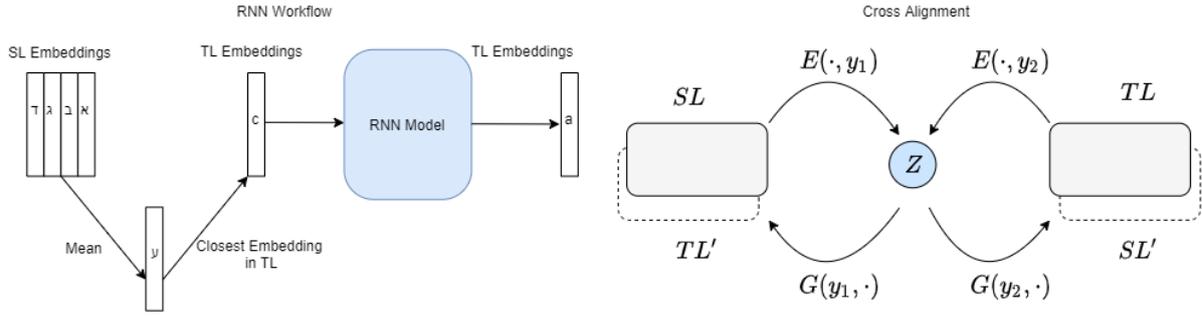

Figure 1: (Left) **RNN workflow.** The mean of the source language word embeddings is taken and the closest match in the target language domain is found. This embedding is then passed as the initial character to the RNN model. (Right) **Cross-alignment** (Shen et al., 2017). A representation of the cross-alignment method. $SL$ and $TL$ are two language domains with different styles $y_1$ and $y_2$. Encoder E maps words to its content representations in the shared latent content space Z. Generator G generates the same words when using the original style. When using a different style it proposes new words by aligning $SL'$ with $TL'$ and vice versa at the distributional level thus realizing style transfer via cross-alignment.

We note a few disadvantages of the above method: (i) Indicator selection: By using the mean of the characters embeddings as an indicator, there is a high chance that multiple source words would have the same embedding resulting in the same words in the target language for different source words, which is clearly undesirable. (ii) Limited dataset: Since our end goal is to extrapolate fictional languages, we have to work with just hundreds of words. Most character level language models, however, work with more examples (on the order of magnitudes higher). These constraints demonstrate the limitations of using a simple RNN.

**Word generator using Seq2Seq.** We use Seq2Seq models for the novel problem of vocabulary generation. The model was trained on a pair of languages using using parallel word translations. We use an encoder-decoder architecture implemented using GRUs (Cho et al., 2014). The input is given as a sequence of characters to the encoder that assigns attention based weight to each character using the attention model in (Bahdanau et al., 2014). This results in the encoder output and hidden state which are inputted to the decoder along with the decoder start token. The decoder then returns a prediction along with the decoder hidden state which is passed back to the model. We use teacher-forcing, where the target character is passed iteratively to the decoder, to make it converge faster and train better.

**Word generator using transformers.** Transformer based language models were also investigated but these performed poorly due to the limited size of the dataset and hence their results have been omitted.

**Word generator using style transfer** We propose using style transfer to let a network learn the underlying style of the languages and act as a vocabulary generating tool. Our training data is not necessarily parallel as the number of words in the English vocabulary is much larger than the few hundred words available in the fictional language. In this case, the network also learns from English words that have no parallel in the fictional language. The arbitrary nature of meaning-to-form mapping (De Saussure, 2011) supports this approach. We use an encoder-generator model following Shen et al. (2017), who proposed achieving style transfer through cross alignment. The encoder takes a word from the source language and encodes it to a style-independent content representation. The generator takes this representation, applies the style of the target language, and then outputs a proposed word belonging to the target language. If the generator uses the same style as the encoder, then the system would act as a variational autoencoder outputting words with the same style as the source language. However, if the generator uses the style of the target language, the system would act as a transfer model generating words similar to those belonging to the target language. The generative part of this network is a GRU. The alignment of the generated latent variables is strengthened by using a cross-aligned structure (Shen et al., 2017). The latent variables are then inputted to a discriminator implemented using a convolutional neural network. The training objective is composed of the reconstruction loss of the generator and the adversarial loss of the discriminator.

**Analysis measures** We propose methods to quantitatively and qualitatively analyze the output words enabling us to gauge network performance.
1. *Qualitative analysis measure.* We would like the words to (i) be on average, around the same length as preexisting words of the target language; and (ii) have the right distribution of vowels so they can be pronounced. A proposed word would be considered successful if it "sounds/feels like it belongs to the language". This requires human evaluation. Note that a complete quantitative analysis (without human intervention) was not possible also for Shen et al. (2017) who used style transfer for sentiment classification.

2. *Quantitative analysis measure.* We propose the following strategy to quantitatively analyze the viability of the output words: instead of training the network on the vocabulary of English and the fictional language, we shall train it on a subset of vocabularies of two 'unrelated' natural languages (e.g., English and Hebrew), where one of them will act as our "fictional language". If the network is able to generate pre-existing Hebrew/English words that it was not exposed to during training, it means that the network has succeeded in extrapolating the language. If the same observation occurs over multiple natural language pairs, we can conclude that the network should also be capable of extending a fictional language. This means that given English and a fictional language, the network will be able to produce new words that the author would have created if they had extended the fictional vocabulary.

**Experiments**

We elaborate on the implementation of our proposed networks including the datasets created and used for training. By using natural languages, we are able to test the viability of our proposed networks. We also quantitatively and qualitatively analyze the results.

To perform vocabulary generation using the Seq2Seq and style transfer networks, datasets with words written in the two different languages are required. During training, the two networks were fed with a pair of language datasets so that they could learn the underlying content representations. Once the networks were trained, we tested each network by inputting words in the source language that the network never saw and asking the network to "translate" the words to the target language.

To measure network performance, we first tested the networks on natural languages. We used multiple language pairs. In this section, we focus on the results for the English-Hebrew pair. We selected the 850 most common English words spoken in the United States (Norman, 2019) and their translations in Hebrew. The networks were then trained over the English-Hebrew language pair. At inference, we simply provided an English word to obtain a "Hebrew translation", and vice versa. By limiting the dataset to hundreds of words, we simulate the approximate size of a fictional language vocabulary.

We provided the trained networks with 100 Hebrew words that were not seen during training and asked the networks to propose English translations. Among the 100 English words proposed by each network, Seq2Seq generated 19 unique words that are part of the English vocabulary (but were not seen during training), while the style transfer network generated 20 such words. The Seq2Seq network also proposed 65 unique words that do not currently belong to the English dictionary, while the style transfer proposed 67. In total, when translating 100 words, the Seq2Seq network proposed 84 new unique English words, while the style transfer network generated 87. These results show that quantitatively, the network performances appear comparable.

We also investigated a simple RNN model which serves as a baseline comparison. While treating English as our target language, the RNN was trained on the 850 Hebrew words that were produced from the (Norman, 2019) dataset. We tested the RNN by giving it 100 Hebrew words and asking it to propose

| English word | Proposed Hebrew translations | | | Hebrew word | Proposed English translations | | |
| --- | --- | --- | --- | --- | --- | --- | --- |
| | RNN | Seq2Seq | S.T. | | RNN | Seq2Seq | S.T. |
| silver | ליפורצ | להתמודד | להבות | להתמודד | hour | ale | seaply |
| thank | טור | משחות | לאין | לשחות | ing | starv | wole |
| branch | זור | צמות | סרגמה | טווח | jout | poper | strice |
| match | חור | מול | משלה | מול | jon | ap | seppare |
| suffix | להיל | חשה | לשתתת | אשה | keat | the | merd |
| especially | ילהת | לא | קעלאוון | נעל | got | gurn | fillice |
| fig | זור | כתף | מי | כתף | joun | wn | rore |
| afraid | וליל | התפוכית | אתרנה | התפשטות | ing | seve | persice |
| huge | טורה | ססר | לשר | לסדר | ing | treat | beat |
| sister | לה | מפשור | ללמתח | מחנה | goter | ent | moun |
| steel | דיר | להמצה | לעות | להמציא | hour | mridy | are |
| discuss | כול | כותנה | התלאל | כותנה | ing | put | tear |
| forward | דיר | נפוד | תתפרה | נולד | hour | bubline | rotunt |
| similar | דיר | תייי | לשמות | לקבוע | keat | wet | beet |
| guide | טור | רבעגורורורורור | מפרה | רבע גלון | gat | win | wilk |
| experience | ילה | תחוע | קערורה | תשע | got | tet | st |
| score | דירה | משומיות | לעור | משאית | hour | twice | at |
| apple | טיר | רעש | אוור | רעש | fine | dide | brant |

Table 1: **Proposed translations using RNN, Seq2Seq, and Style Tranfer (S.T.) methods.** (Column I) English words we want translated. Proposed Hebrew translations using (column II) RNN; (column III) Seq2Seq network; (column IV) Style Transfer network. (Column V) Hebrew words we want translated. Proposed English translations using (column VI) RNN; (column VII) Seq2Seq network; (column VIII) Style Transfer network. Words marked in green are preexisting English words that were generated but the network saw during training. Words marked in blue are preexisting English words that were generated but the network never saw. Words marked in red are generated words that occur more than once in the column.

English translations. The dictionary mapping that enables this translation mechanism is described in the previous section. An analysis of the output showed that out of the 100 proposed English words, the network was able to generate only 12 words that already belong to the English language (but were not seen during training). The network also produced only 31 new unique words that currently do not exist in the English language. This shows that the RNN performance is quite limited when compared to the style transfer and Seq2Seq networks.

Table 1 displays English words with the proposed Hebrew translations and Hebrew words with their proposed English translations using the three networks. By analyzing the words generated, one can note that the Style Transfer and Seq2Seq networks display a wide range in character usage, word length, and have no repetitions. The RNN however performs relatively poorly demonstrating the need for the more advanced solutions we use. Notice that the RNN proposed a few English words that exist in the language, however, the network had already seen these words during training so these cannot be considered as novel generated words. Such words, that exist in the language, but the networks already saw, are marked in green in Table 1. Red color marks repetitions of words that have already occurred in the table and blue color marks preexisting English words that were generated but the network never saw. The existence of blue words show that the networks were able to extrapolate the English vocabulary from a limited subset. Also note that most of the words marked in black, which are proposed words that do not currently exist in the English vocabulary can be pronounced and do sound and feel like they could belong to the English language. A qualitative analysis to study the same is performed through a user study described in the next section. It is also interesting to note that all networks generated words that are short and have vowels - qualities that are both desirable.

## Discussion

Following the evaluation of our proposed strategy above, we make further in-depth investigations into the results produced by our network. We do this by conducting a user study, looking for linguistic connections between words and their proposed translations, and by using the network to extend the vocabulary of the fictional "Ancient Language".

| English word we want translated | Proposed translations | | | | |
|---|---|---|---|---|---|
| | Arabic | Hindi | Spanish | Amharic | Russian |
| silver | شرة | विड़का | silverian | ሆንባት | пресносать |
| thank | الصادة | पाड़ी | thanak | ራሩ | строл |
| branch | ذرس | पिरत्ध | brícancak | ተነገም | попрастить |
| match | قائرة | पूला | macta | ቅር | прость |
| suffix | متطفة | परसिजक | suficar | መሰሰለም | реститьс |
| especially | عيدية | कालापनीकीजिये | sespriclama | እንተለቀል | нострммемнать |
| fig | موق | सिक | figa | ኦባ | меть |
| afraid | ماوق | परतर | faricadiar | ተገሰኝ | постедиться |
| huge | جل | प्या | guego | ደጠሻ | вель |
| sister | حر | हिलाना | sistrecian | እንፈዳም | ресносться |
| steel | قطيدة | हाज़ी | sentelo | ንግል | норани |
| discuss | تعرد | मेप्रनाने | discursina | አለረን | перурустый |
| forward | كبمعة | सेतरत | forricadar | አንገት | поставиться |
| similar | مواع | नालकी | similarian | አንገድል | престостени |
| guide | عيم | मुक्य | guedido | አዋታ | дежить |
| experience | يصار | काप्ताकिनिके | experciende | ኤያንተለ | носторомный |
| score | المن | खिबा | bisación | እንኑት | проза |
| apple | المينة | पूरका | sprepadía | ተቀሳል | порона |

Table 2: **Proposed translations.** English words and their proposed translations to Arabic, Hindi, Spanish, Amharic, and Russian. Note that a different trained network was used for each target language.

| Network | Dataset E-H words | Quantiative analysis | | | Qualitative analysis | |
|---|---|---|---|---|---|---|
| | | EWNS | PNW | TUW | Mean | Median |
| Seq2Seq | 200-200 | 145 | 668 | 813 | 2.6 | 3 |
| Style transfer | 200-8000 | 184 | 842 | 1026 | 3.1 | 3 |
| Seq2Seq | 850-850 | 217 | 740 | 957 | 3.4 | 4 |
| Style transfer | 850-8000 | 191 | 895 | 1086 | 3.3 | 3 |

Table 3: **Quantitative and Qualitative analysis.** (Column I) Networks used for word generation. (Column II) The English-Hebrew dataset that the network was trained on. (Column III) EWNS - Generated preexisting English Words that were Not Seen during training. (Column IV) PNW - Proposed Novel Words currently not in the English vocabulary. (Column V) TUW - Total Unique Words generated. (Column VI) Mean score (out of 5) from user study. (Column VII) Median score (out of 5) from user study.

**User Study.** For the user study, we generated words from the Seq2Seq and style transfer networks trained on different datasets (see table 3) and asked judges to rate them from 0 (cannot be an English word) to 5 (definitely an English word). Each network trained on the different datasets as seen in table 3 was asked to generate 1300 new words. We conducted a quantitative analysis on the generated words by analyzing the number of preexisting English words and the total number of unique words that the network could generate. The results for the same can be viewed in table 3. It can be seen that the style transfer network for both datasets is able to generate more unique words, however the Seq2Seq network trained on the 850 word language pair performs slightly better while generating preexisting English words.

To analyze the quality of the output words, 48 judges took part in the user study and a total of 2400 words were rated- 600 each corresponding to the rows in table 3. Qualitatively, the Seq2Seq performs comparably well to the style transfer network when trained on the dataset with 850 English words. However, both quantitatively and qualitatively, style transfer outperforms Seq2Seq when it comes to the smaller dataset containing 200 English words. This is because the style transfer network is able to take advantage of the unpaired style of training unlike the Seq2Seq network. Since the size of fictional datasets are largely limited, using the style transfer network should be preferable.

| English word we want translated | Proposed Hebrew translation |
|---|---|
| block | סוין |
| blockade | סובינון |
| throw | להיפור |
| threw | להיפר |
| throwing | להיינור |
| like | נובה |
| likely | נסובה |
| hear | לקרון |
| heard | להראון |
| hearing | להרגמי |
| task | מולה |
| tasking | למימובי |
| wise | נפר |
| wisdom | נופה |
| imagine | מאפרים |
| imagined | מהגאית |
| imagination | מהלאימון |

Table 4: **Linguistic connections.** English words originating from the same root and their proposed Hebrew translations.

| English word we want translated | Proposed Ancient language translation |
|---|---|
| understand | dunendarvald |
| understood | duendorva |
| understanding | dunendanvaldrina |
| fiction | fitingoting |
| fictional | fitingatilona |
| revolt | relthilg |
| revolution | reluthingo |
| revolutionary | eruthilnadorin |
| frown | frongh |
| frowned | ffrendhr |
| frowning | frönnghing |
| imagine | minandae |
| imagined | mindanaedí |
| imagination | mingainatoind |
| generate | engaerth |
| generated | geengardethin |
| generation | gengarinona |
| generating | gengarnathin |

Table 5: **Linguistic connections.** English words originating from the same root and their proposed Ancient language translations.

**Testing Style Transfer network on multiple languages.** In addition to the in-depth analysis for the Hebrew-English pair, we also conducted multiple experiments with our preferred style transfer network for different language pairs. For the convenience of readers around the world, we provide in Table 2 English words along with our network's proposed novel generated words (proposed translations) in Arabic, Hindi, Spanish, Amharic, and Russian. In the two languages that we understand in the Table, we observe the same resemblance to the target language that we had with the generation of words in English and Hebrew. We believe that the readers who understand the languages will observe the same when going through the results in Table 2. All these results demonstrate the capability of our proposed strategy in creating novel words in a given language that contains only a small number of existing words.

**Linguistic connections.** We also investigated linguistic connections between the translations of words from the same root. We selected English words originating from the same root and asked the trained style transfer network to propose Hebrew translations. Table 4 shows English words and their proposed translations. As can been seen in the table, the proposed Hebrew translations also share common word roots which strengthens our proposition of using style transfer for novel word generation. It is very interesting that the network was able to achieve this despite the non-parallel nature of training and the small number of words used from target language.

We also investigated linguistic connections between the proposed translations of words from the same root. Table 5 shows a few examples of English words belonging to the same root along with the proposed translations. Notice how the proposed translations also have common word roots, which corroborates our hypothesis.

**Ancient Language.** Additional examples of words generated for the Ancient Language by the style transfer network can be viewed at table 6. These examples were generated by training on a dataset of 400 Ancient Language words and 8000

| Christopher Paolini's definitions | | Our network's proposals | | | |
| --- | --- | --- | --- | --- | --- |
| English | Ancient Language | English | Ancient Language | English | Ancient Language |
| create | aldanarí | artisan | rathans | torch | throch |
| elf | älfa | historian | istharina | constellation | nesthalona |
| halt | blöthr | modest | medts | racism | varsia |
| reduce | brakka | pony | ponn | forbid | frodhi |
| honor | celöbra | grimace | gramia | feign | feing |
| sage | chetowä | redundancy | renduvandí | slant | skant |
| brother | darmthrell | decisive | desviverna | ambition | maithinor |
| mists | datia | abundant | bandanthar | coerce | ceror |
| grow | eldhrimner | teenager | ternanga | aisle | iales |
| invoke | ethgrí | harsh | harsh | patience | vaethinva |
| marked | fódhr | fragment | franthandí | fisherman | frishvandr |
| sing | fyrn | pigeon | gieno | bathtub | bathhr |
| chant | gala | soprano | sorana | lend | lend |
| luck | guliä | freckle | frelka | rubbish | brisha |
| height | haedh | curl | völrr | shareholder | sharelhra |
| whale | hwal | continuation | onvithanda | decay | deyja |
| my | iet | housewife | shvaedhrin | deter | deetr |
| truth | ilumëo | stir | stirr | feminine | ferininve |

Table 6: **Ancient language.** English words and their translation into the Ancient language. Left: Words that Christopher Paolini defined. Middle&Right: Our network's proposals for words that were never defined.

English words. We could use more English words since we could exploit the nature of non-parallel dataset of the style-transfer network to further improve the results.

**Using non-parallel data.** It is worth noting that by using the non-parallel training approach, we are not giving up linguistic information provided by parallel links. As shown by the linguist De Saussure (2011), meaning-to-form mapping is almost entirely arbitrary and there is no simple or logical relation between a certain word and its translation (for languages that are not phonologically related). We can also use this property (deferred to future work) to increase the size of the fictional vocabulary by using examples from the language that the fictional language is based on. For example, the 'Ancient Language' from the Inheritance Cycle is based on ancient Norse and Celtic. By incorporating words from these languages, the number of examples that the network sees during training can be increased significantly. The only constraints at play are the phonological and morphological rules of a language. By using style transfer, the aim is for the network to learn these phonological and morphological rules (the underlying style) which we believe that the network was able to achieve.

## Conclusion

We proposed a deep learning based approach to enable the completion of fictional language vocabularies. We showed how style transfer can be used as a vocabulary generation tool; proposed ways to analyze the outputs; and successfully demonstrated the extension of a fictional vocabulary.

We investigated RNN, Seq2Seq, style transfer, and transformer models and quantitatively and qualitatively analyzed their results. We found that the style transfer network had the best performance because of the limited nature of the training dataset. While we used the style transfer technique from Shen et al. (2017) to perform the novel words generation, one may use other textual style transfer frameworks in a similar way (Zhao et al., 2017; Zhang et al., 2018; Li et al., 2019).

Apart from the obvious applications in the entertainment industry, our proposed methodology could also be relevant for translating modern words such as internet, modem, etc, to existing languages in an automated way that takes into account the style of the target language. It could also be interesting to use the network to help revive low-resource and endangered natural languages.